\documentclass[runningheads]{llncs}
\usepackage{graphicx}

\usepackage{subfigure}

\usepackage{hyperref}

\usepackage{url}

\usepackage{ulem}

\usepackage[utf8]{inputenc} 
\usepackage[T1]{fontenc}    

\usepackage{amsmath,amsfonts,amssymb,amsthm,dsfont}
\usepackage{booktabs}
\usepackage{float}
\usepackage{wrapfig} 
\usepackage{multirow}
\usepackage{color}
\usepackage{soul}

\def\R{\mathbb{R}}

\usepackage{color}
\usepackage{todonotes}

\title{Hypernetwork functional image representation}
\author{Sylwester Klocek \and
\L{}ukasz Maziarka \and
Maciej Wo\l{}czyk \and
Jacek Tabor \and
Jakub Nowak \and
Marek \'Smieja }
\authorrunning{S. Klocek et al.}
\institute{Faculty of Mathematics and Computer Science\\
Jagiellonian University\\
\L{}ojasiewicza 6, 30-348, Krak\'ow, Poland\\
\email{marek.smieja@ii.uj.edu.pl}}

\begin{document}

\maketitle
\begin{abstract}
Motivated by the human way of memorizing images we introduce their functional representation, where an image is represented by a neural network. For this purpose, we construct a hypernetwork which takes an image and returns weights to the target network, which maps point from the plane (representing positions of the pixel) into its corresponding color in the image. Since the obtained representation is continuous, one can easily inspect the image at various resolutions and perform on it arbitrary continuous operations. Moreover, by inspecting interpolations we show that such representation has some properties characteristic to generative models. 
To evaluate the proposed mechanism experimentally, we apply it to image super-resolution problem. Despite using a single model for various scaling factors, we obtained results comparable to existing super-resolution methods. 

\end{abstract}

\section{Introduction}

Classical machine learning approaches are based on optimizing a predefined class of functions, such as linear or kernel classifier \cite{scholkopf2001learning}, Gaussian clustering \cite{banfield1993model}, least squares regression \cite{geladi1986partial}, etc.. With the emergence of deep learning, we learned how employ general nonlinear functions given by complex neural networks. Replacing shallow methods by deep neural networks opened new opportunities in machine learning, because arbitrarily complex functions can be approximated \cite{bengio2015deep}.

However, when one considers a typical representation of the data, we still use a somehow shallow and restrictive vector approach. Although real data, such as sound or image, have analog character, we represent them in an artificial vector form. In consequence, one cannot easily access an arbitrary position of the image, rescale the image or perform (even linear) operations like the rotation \cite{SRCNN-2016} \cite{lin2017feature}, \cite{danelljan2016beyond} without the additional use of interpolation \cite{gao2011bilinear}.  Concluding, it is impossible to satisfactory utilize advanced deep learning methods without creating natural mechanisms for data representation. 

The aim of this paper is a proof of concept that one can effectively construct and train {\it functional representations} of images. By a {\it functional (or deep) representation of an image} we understand a function (neural network) $\mathcal{I}:\R^2 \to \R^3$ which given a point (with arbitrary coordinates) $(x,y)$ in the plane returns the point in $[0,1]^3$ representing the RGB values of the color of the image at $(x,y)$. Observe that given a functional image representation, we can easily obtain a corresponding vector representation by taking $(\mathcal{I}(i,j))_{i=1..k,j=1..n}$. In constrast to the vector form, the functional image representation in its idea can be compared by the way a human represents the image\footnote{we can reasonably hypothesise that a human representation of an image in the memory is given by some neural network}.

\begin{figure}[t]
    \centering
    \includegraphics[width=0.7\textwidth]{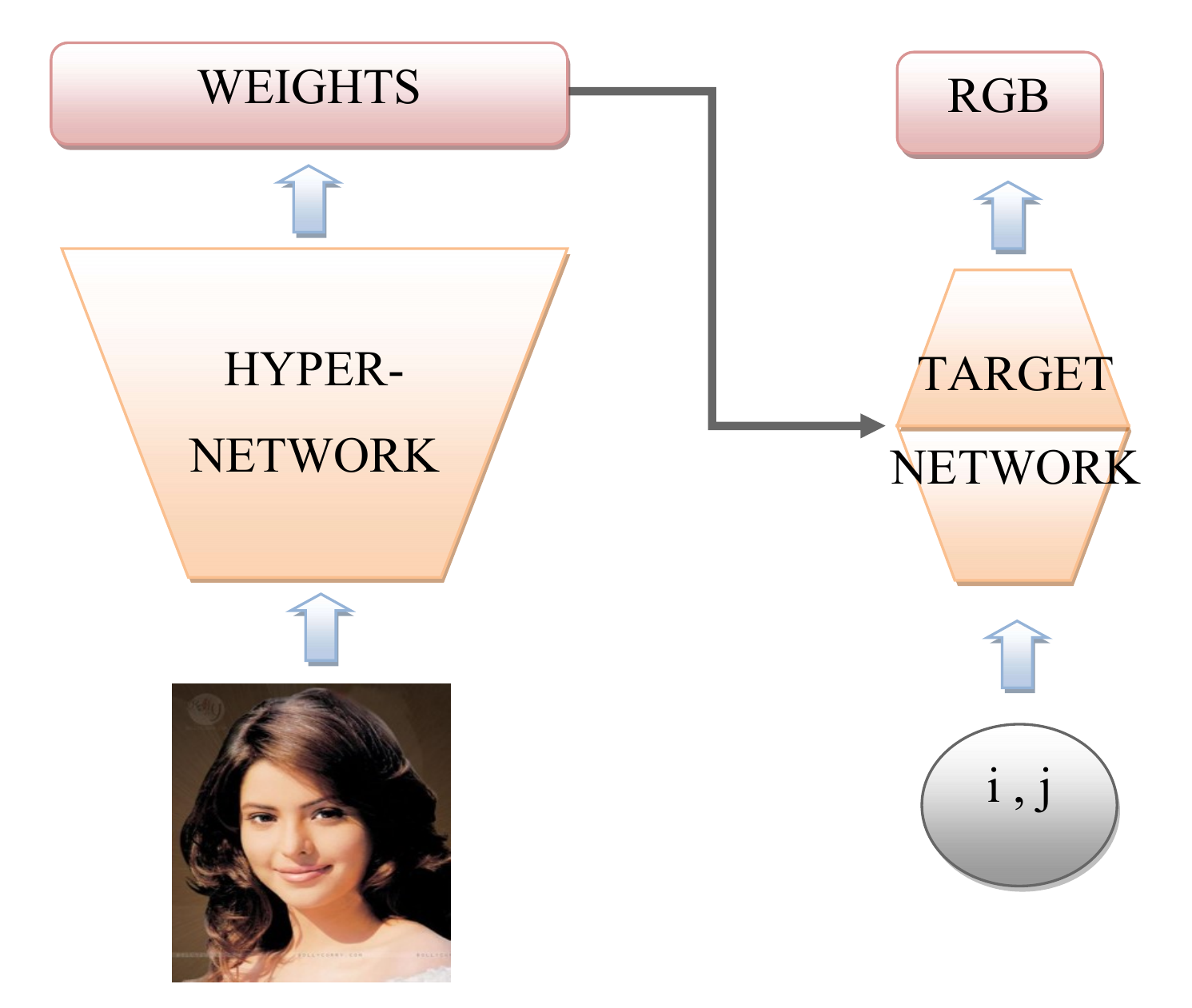} 
  \caption{The scheme of our approach. The hypernetwork takes an image and produces the weights to target network, which is responsible for approximating an image at every real-valued coordinate pair $(i,j) \in [0,1]^2$.\label{fig:idea}} 
\end{figure}

The main achievement of the paper is the functional image representation constructed with the use of hypernetwork \cite{hypernetworks-2016}. The hypernetwork takes an image and produces a target neural network, which is an approximation of the input image, see Figure \ref{fig:idea} for illustration. Instead of creating the whole architecture of the target network from scratch, the hypernetwork returns only the weights to its fixed, predefined architecture. This allows us to effectively train both hypernetwork and target networks at the same time, using stochastic gradient descent.

\begin{figure}[t]
    \centering
    \includegraphics[width=\textwidth]{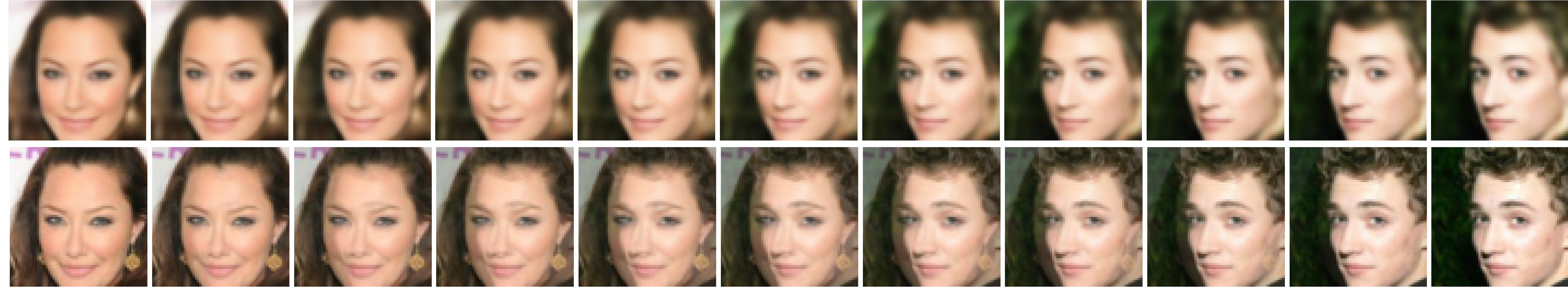} 
  \caption{A linear interpolation between weights of two target networks (upper row) compared with a typical pixel-wise interpolation (bottom row). Images produced by target networks with interpolated weights corresponds to natural images coming from true data distributions. There in no superimposition of images, unlike in the case of pixel-wise interpolation. \label{fig:celeba_interpolation}} 
  
\end{figure}

\begin{figure}[t]
    \centering
    \includegraphics[height=3.7cm]{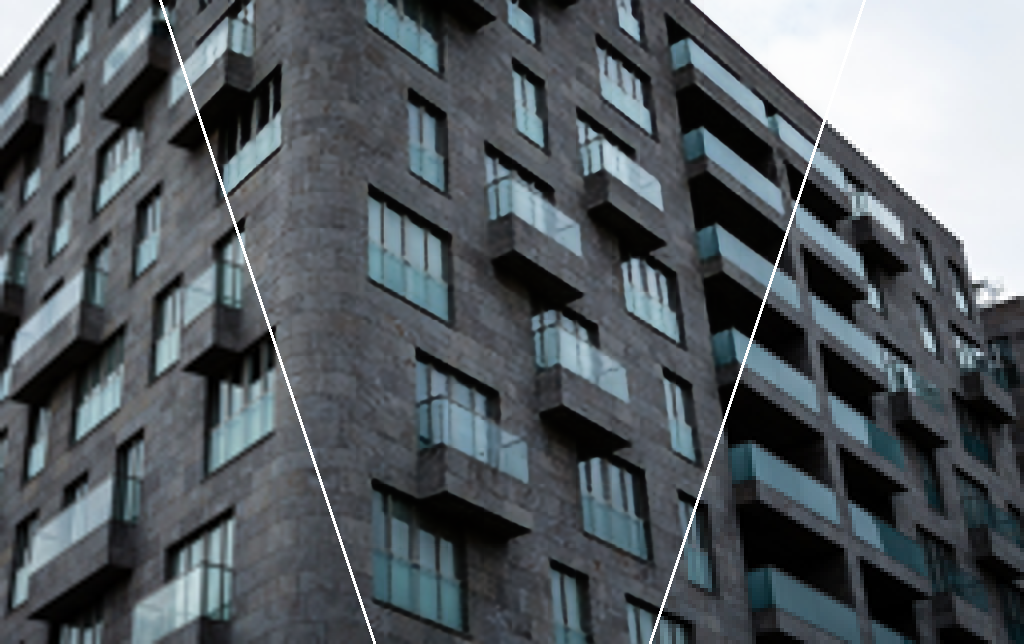} \quad
    \includegraphics[height=3.7cm]{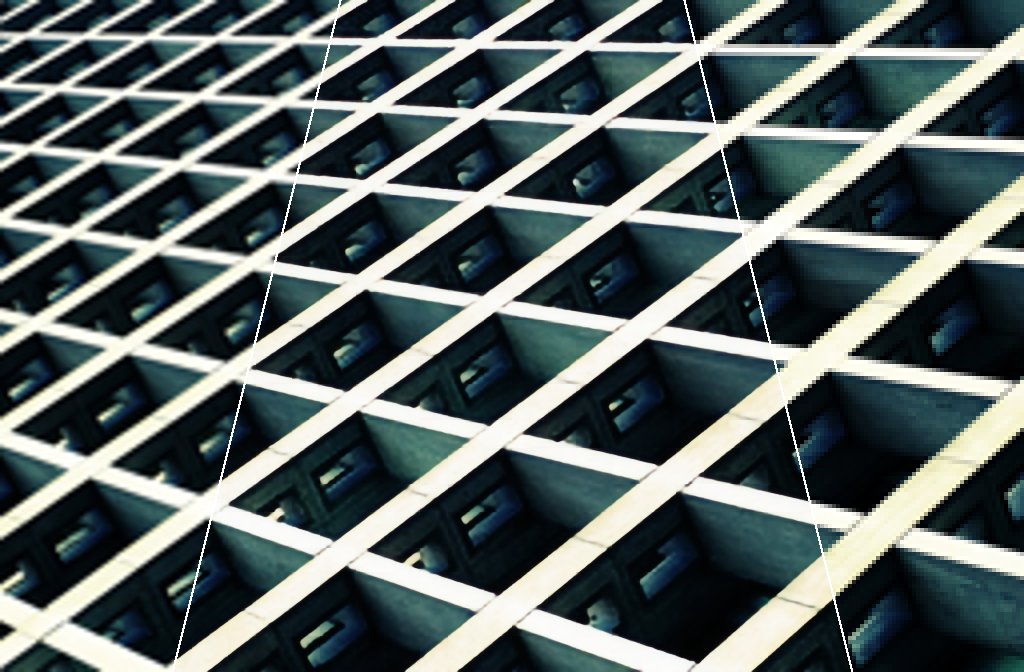}
    \caption{Higher resolution images (scaling factor $\times 4$) obtained with a use of the bicubic interpolation (left) and our method (middle). A low-resolution input image is on the right side.\label{fig:img1}}
\end{figure}

\begin{figure}[t]
\centering
\includegraphics[width=0.7\textwidth]{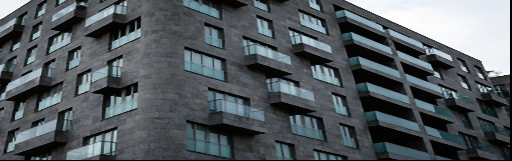}
\caption{Resizing the image to non-standard $2.5 \times 1$ resolutions. \label{fig:img2}}
\end{figure}

We summarize the advantages and applications of our representation:
\begin{enumerate}
    \item We use a single hypernetwork, which gives a recipe of how to represent images by target networks. In consequence, we return the functional representation of any input image at test time without additional training. 
    \item It is a well-known fact that true images represent a manifold embedded in high-dimensional space \cite{yeh2017semantic}. In consequence, it is difficult to interpolate between two images in the pixel space and not to fall out of the distribution of true  images. In contrast, the geometry of target networks' weights is less complex (Section \ref{sec:interpolation}). We have verified that simple linear interpolation between weights of target networks representing images leads to natural images (Figure \ref{fig:celeba_interpolation}).
    \item Due to the continuity of this representation, we are not limited to a fixed resolution, but can operate on a continuous range of coordinates. To confirm this property, we applied our approach to a  super-resolution task (Figure \ref{fig:img1}). In contrast to typical super-resolution methods \cite{zhang2018residual}, \cite{SRCNN-2016}, which are trained for a single scaling factor, our model is able to upscale the image to any size. While its effects are competitive compared to existing approaches (Section \ref{sec:superres}), we can also use non-standard scales at test time (Figure \ref{fig:img2}).
\end{enumerate}

\section{Related work}

We briefly outline related approaches. We start with describing the hypernetwork model. Next, we discuss typical methods for image representation. Finally, we move on to super-resolution techniques.

Hypernetworks were introduced in \cite{hypernetworks-2016} to refer to a network generating weights for a target network solving a specific task. The authors aimed to reduce the number of trainable parameters by designing a hypernetwork with a smaller number of parameters than the target network being generated. Making an analogy between hypernetworks and generative models, the authors of \cite{sheikh2017stochastic}, used this mechanism to generate a diverse set of target networks approximating the same function. This is slightly similar to our technique, but instead of creating multiple networks for the same task, we aim at generating individual network for each task (image). In \cite{sheikh2017stochastic}, the hypernetwork was used to directly maximize the conditional likelihood of target variables given a certain input. Moreover, hypernetworks were also applied in Bayesian context \cite{krueger2017bayesian}, \cite{louizos2017multiplicative}. Finally, the authors of \cite{SMASH-2017}, \cite{GHS-2018}, \cite{stochastic-2018} used the hypernetwork mechanism to create or improve the search of the whole network architecture. 

Images are commonly represented as two-dimensional matrices with a fixed size (resolution). Due to the high redundancy in image features, one can use auto-encoders to create compressed representation in lower-dimensional space \cite{baldi2012autoencoders}. By adding 
controlled noise to input data at training stage, we can obtain a representation, which is less sensitive to the image perturbations at test stage \cite{vincent2008extracting}. Auto-encoders can also be used as a basis for generative models, such as VAE \cite{kingma2013auto} or WAE \cite{tolstikhin2017wasserstein}, which can learn an even more compact representations and generate new images using a decoder network \cite{liu2017unsupervised}. Although auto-encoders can be used to transfer knowledge to less explored domains \cite{wang2013learning}, the representations they learn are in a vector space. To represent an image as a function, one can approximate pixel intensities by a regression function, e.g. polynomial or kernel regression \cite{takeda2006kernel}, or interpolate between neighbor pixels, e.g. bicubic or B-spline interpolation \cite{hwang2004adaptive}, \cite{gao2011bilinear}, \cite{unser1991fast}. More advanced approaches rely on using wavelets \cite{lee1996image} or ridgelets \cite{do2003finite}, which play a key role in JPEG2000 compressor \cite{christopoulos2000jpeg2000}. Nevertheless, the aforementioned methods require manual selection of the regression model, which makes them difficult to use in practice. To overcome these problems we follow deep learning approach and allow the hypernetwork to select the optimal function based on the data set.

Super-resolution area has been dominated by deep learning models. One of the earliest approaches applied a lightweight convolutional neural network to directly map a low-resolution image to its high-resolution counterpart \cite{SRCNN-2016}. In contrast, the authors of \cite{kim2016accurate} used a very deep convolutional network inspired by the VGG Net to predict residuals instead of the output image itself. The paper \cite{ledig2017photo} introduced a discriminator network as in the case of GANs to make upscaled images more realistic. In \cite{lim2017enhanced} the authors reused residual networks, but removed unnecessary modules, which resulted in increasing the speed and model stabilization. The authors of \cite{zhang2018residual} also used residual networks, but exploited the hierarchical features from all the convolutional layers instead of the last one. Despite a huge progress in super-resolution area, most of existing models are trained for a single scale factor. In contrast, the proposed hypernetwork technique allows us to upscale the image to multiple sizes using only a single hypernetwork.

\section{Functional image representation}

In this section, we introduce our functional image representation. First, we describe our learning model and define its cost function. Next, we discuss the architectures of the hypernetwork and the target network.

\subsection{Hypernetwork model}

Let $f:[0,1]^2 \to [0,255]^3$ be a function describing the image. In practice, we only observe pixel intensities in a fixed grid. To improve this discrete representation, we aim at creating a function:
$$
T_\theta(i,j) = T((i,j),\theta): [0,1]^2 \times \Theta \to [0,255]^3,
$$
which approximates RGB values of each coordinate pair $(i,j) \in [0,1]^2$. Our objective is to find an optimal weight vector $\theta \in \Theta$ for every image.

In the simplest case, $T_\theta$ can be obtained by linear or quadratic interpolation \cite{gao2011bilinear}, which however may not be sufficient for approximating complex image structures. To achieve higher flexibility, one could model every image with a specific neural network $T_\theta$. Nevertheless, training separate networks for each image using backpropagation may be computationally inefficient. 

We approach this task by introducing a hypernetwork
$$
H_\varphi: X \ni x \rightarrow \theta \in \Theta, 
$$
which for an image $x \in X$ returns weights $\theta$ to the corresponding target network $T_\theta$. Thus, an image $x$ is represented by a function $T((i,j);H_\varphi(x))$, which for any coordinates $(i,j) \in [0,1]^2$ returns corresponding RGB intensities of the image $x \in X$.

To use the above model, we need to train the weights $\varphi$ of the hypernetwork. For this purpose, we minimize classical mean squared error (MSE) over training images. More precisely, we take an input image $x \in X$ and pass it to $H_\varphi$. The hypernetwork returns weights $\theta$ to target network $T_\theta$. Next, the input image $x$ is compared with the output from target network $T_\theta$ over known pixels. In other words, we minimize the expected MSE over the training set of images:
$$
MSE = \sum_{x \in X} \sum_{(i,j)} [x[i,j] - T((i,j); H_\varphi(x))]^2.
$$

Observe that we only train a single neural model (hypernetwork), which allows us to produce a great variety of functions at test time. In consequence, we might expect that target networks for similar images will be similar (see experimental section). In contrast, if we created a single network for every image, such identification would be misleading.

\begin{figure}[t]
\centering
\includegraphics[width=\textwidth]{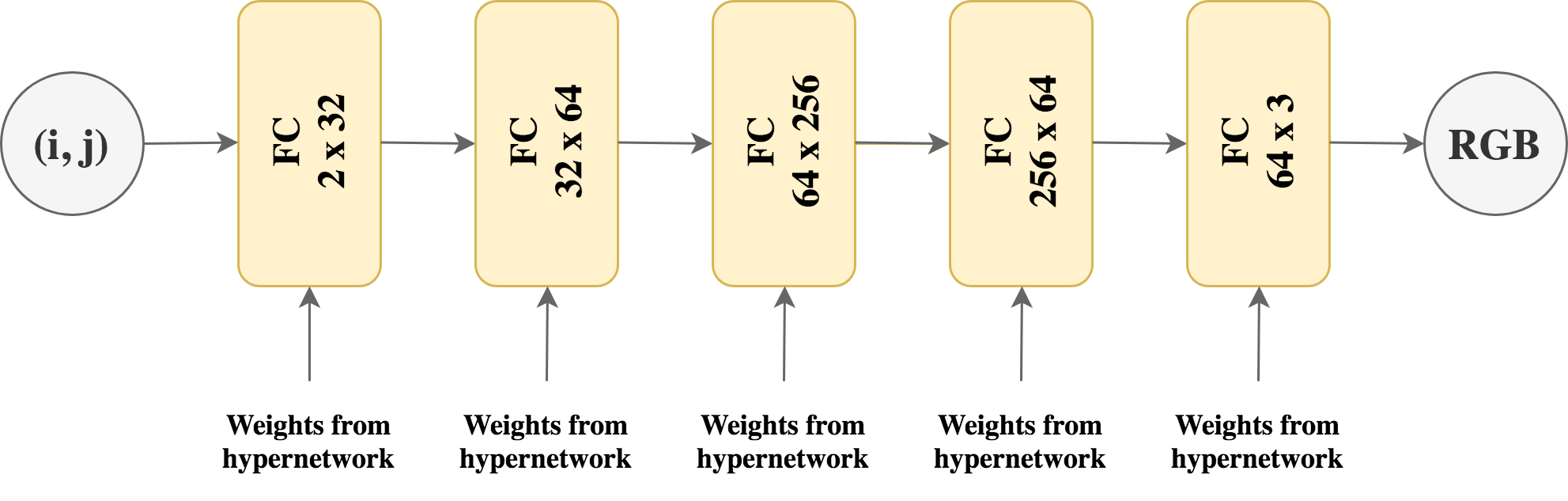}
\caption{The architecture of the target network.}
\label{fig:target-arch}
\end{figure}

\begin{figure}[hp]
    \centering
    \includegraphics[width=\textwidth]{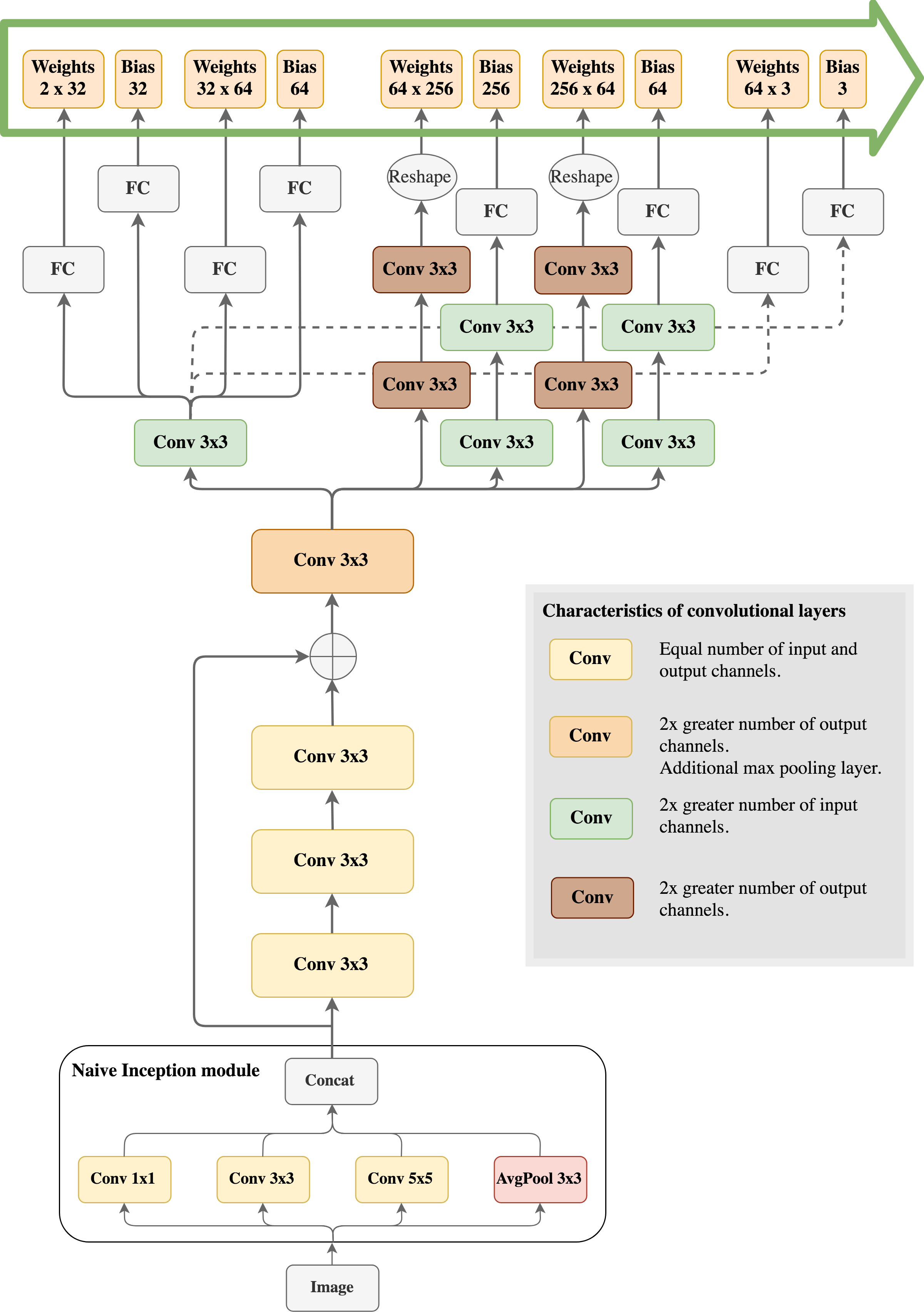}
    \caption{The architecture of the hypernetwork.}
    \label{fig:hypernetwork-arch}
\end{figure}

\subsection{Architecture}

In this part, we present the architectures used for creating functional image representation.

\paragraph{Target network.} An architecture of the target network is supposed to be simple and small. This allows us to keep the performance of training phase at the highest possible level as the target network is not directly trained. Moreover, small networks can be easily reused for other applications.

Target network maps a pair of two coordinates to corresponding three dimensional RGB intensities. The target network consists of five fully-connected layers, see Figure \ref{fig:target-arch} with biases added in each of them. The layers' dimensions are being gradually increased. This is happening up to the middle layer. Later on, they are being decreased. This is because steep transitions of layers' dimensions negatively affect the learning ability of neural network. The matrices used in the target network have the following dimensions:
2x32, 32x64, 64x256, 256x64, 64x3. Additionally, batch normalization is used after each layer \cite{ioffe2015batch}. We have chosen cosine to be the activation function between two consecutive layers. This choice is motivated by a typical approach used in mathematical image transformation, which is based on discrete cosine transform (DCT), see JPEG compression. For our purposes bounded cosine function worked much better than ReLU, which can give arbitrary high outputs\footnote{Other experimental studies report that there are not much difference between using cosine and ReLU as activity function \cite{goodfellow2016deep}.}. We use sigmoid as the activation function for the last layer. No convolutions were used, because the input of the target network is too simple.

\paragraph{Hypernetwork.} 

The hypernetwork used in our model is a convolutional neural network with one residual connection, see Figure \ref{fig:hypernetwork-arch}. There are two main parts of the hypernetwork’s architecture. The first one is common and takes part in generating weights for all of the target network’s layers. Second part, on the other hand, contains several branches. Each branch calculates weights for a different layer of target network. This approach enabled faster training, compared to creating a separate hypernetwork for each layer of the target network.

The task of common layers is to extract meaningful features. This extraction is performed using Naive Inception Module \cite{szegedy2015going} followed by four convolution layers. Inception module leverages three different convolutions and average pooling. It improves network accuracy and does not negatively influence training time. At the very end of common part, we introduced max pooling to decrease the number of features.

Next, there are multiple branches responsible for converting extracted features into actual weights of target network. Dimensions of the target network’s layers vary. Therefore, convolutions with different number of input and output channels are used in the branches. Their role is to adjust the sizes of tensors. Shapes of tensors also need to be modified, hence convolutions are followed by either fully-connected layers or simple reshapes. Fully-connected layers are used in branches generating weights for smaller layers of target network. Batch normalization is used after each layer of hypernetwork and ReLU is chosen as the activation function. 

Finally, to accelerate training phase even more, we reduced number of trainable parameters by adapting an approach from inception network \cite{szegedy2016wojna}. More precisely, we replaced each $n \times n$ convolution with $1 \times n$ convolution followed by a $n \times 1$ convolution.

\section{Experiments}

In this section, we examine the proposed functional image representation. First, we analyze the space of target networks' weights and show some interesting geometrical properties. Next, we use the continuity of our representation and apply it to super-resolution.

\subsection{Target networks geometry} \label{sec:interpolation}

\begin{figure}[hp]
    \centering
  \subfigure[Interpolation between target networks weights.]{\includegraphics[width=\textwidth]{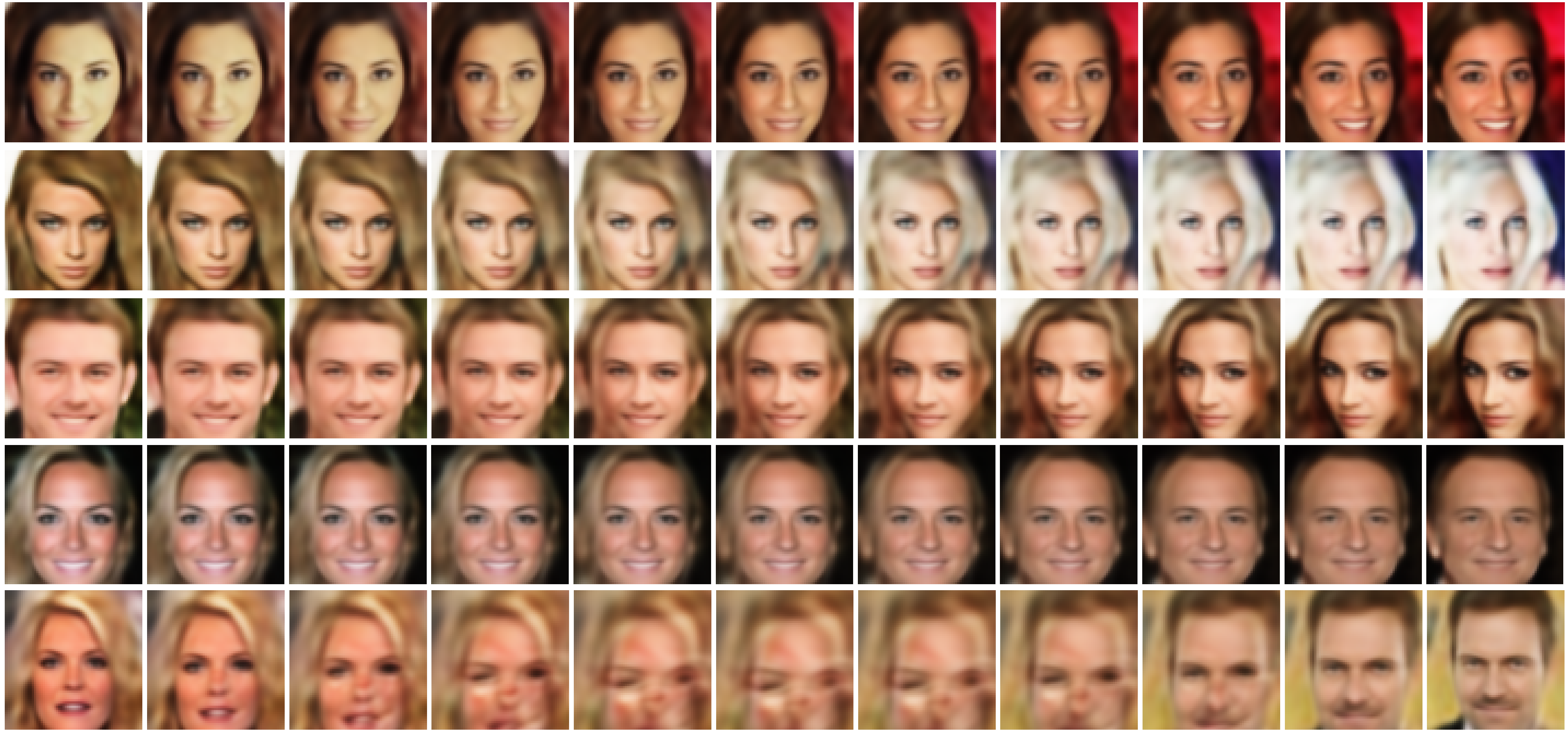} } 
  \subfigure[Pixel-wise interpolation.]{\includegraphics[width=\textwidth]{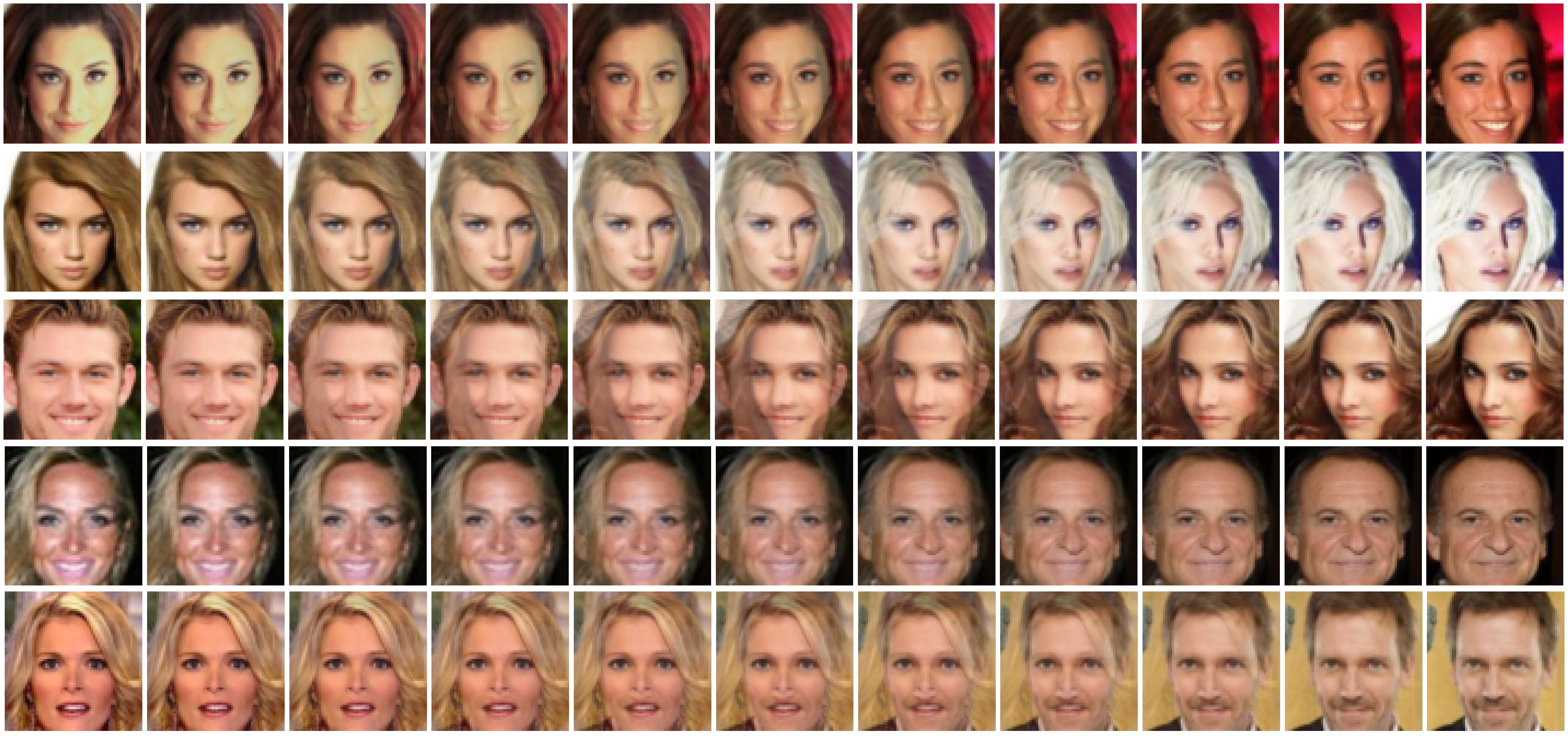}}
  \caption{Comparison of the interpolation between weights of two target networks (a) and linear pixel-wise interpolation (b). We also include an example of unsuccessful interpolation (last row), where interpolation went beyond the manifold of true images.  \label{fig:celeba_interesting_interpolation}} 
  
\end{figure}

\begin{figure}[t]
\centering
  \includegraphics[width=\textwidth]{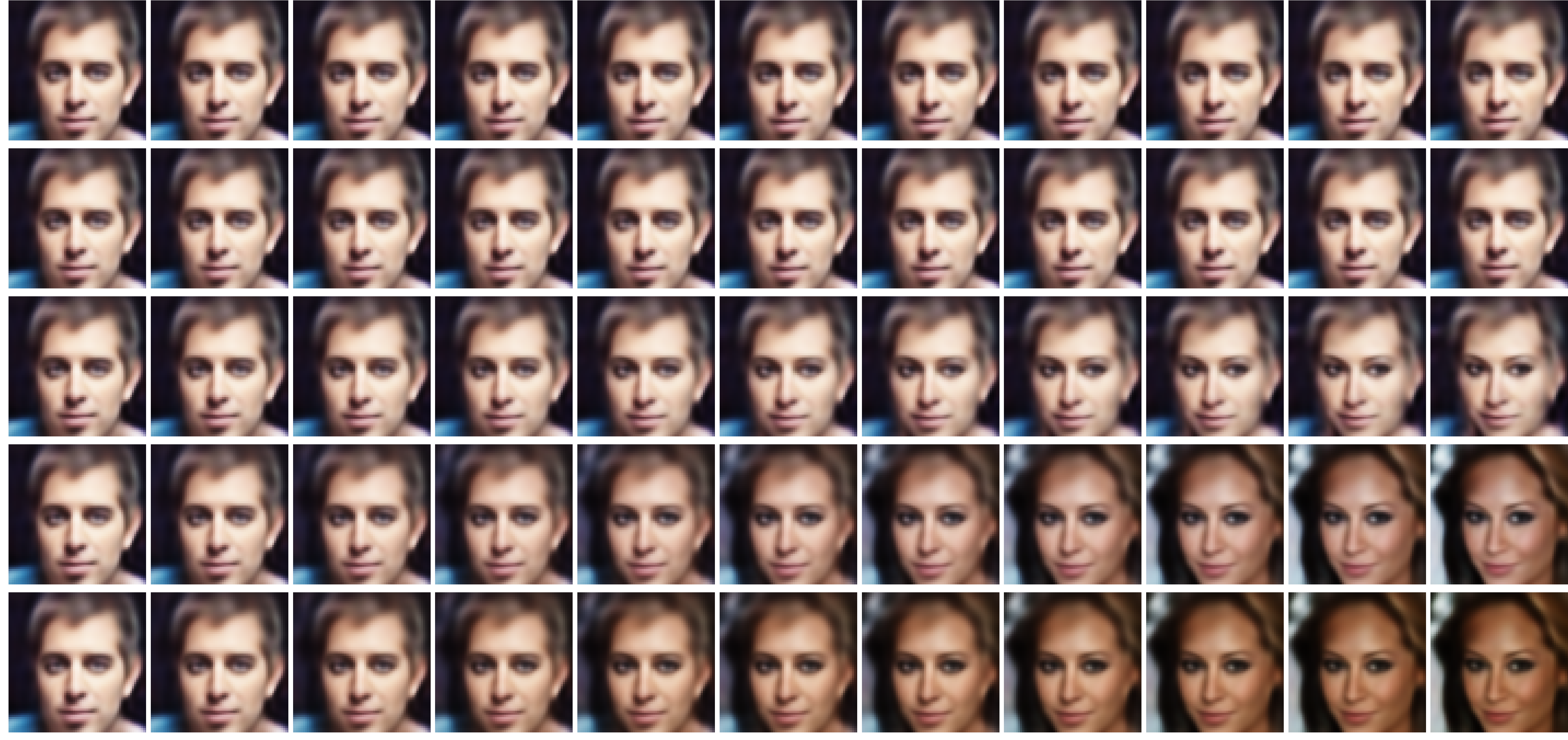}
  \caption{Layerwise interpolation on the CelebA. In $i$-th row we interpolate only over the first  $i$ layers of the network.
  }
  \label{fig:celeba_layerwise}
\end{figure}

It is believed that high dimensional data, e.g. images, is embedded in low-dimensional manifolds \cite{goodfellow2016deep}. In consequence, direct linear interpolation between images does not produce pictures from true data distributions. 

In this experiment, we would like to inspect the space of target network weights. In particular, we verify whether linear interpolation between weights of two target networks produces true images.
We use the CelebA data set \cite{celeba-2015} with trivial preprocessing of cropping central 128x128 pixels from the image and resizing it to 64x64 pixels.

In test phase, we generate target networks for two images and linearly interpolate between weights of these networks. Figure \ref{fig:celeba_interesting_interpolation}a presents exemplary images returned by interpolated target networks. In most cases, interpolation produces natural images which come from the true data distribution (first four rows). It means that we transformed a manifold of images into a more compact structure, where linear interpolation can be applied. It allows us to suspect that similar images have similar weights in their corresponding target networks. Occasionally, interpolated weights produce images from outside of images manifold (last row). This negative effect sometimes occurred when interpolated images were very different, e.g. in this examples, we have a pair of images of people of different sex, hairstyle, skin color and photographed in different poses. These rare cases are accepted, because we did not use any additional constraints. For a comparison, we generate classical pixel-wise interpolation between analogical examples. As can be seen in Figure \ref{fig:celeba_interesting_interpolation}b, the results are much worse, resulting in a superimposition of images.

Going further, we verify a layer-wise interpolation. Namely, we take weights to one target network and gradually change weights of the first $i$ layers in the direction of corresponding weights in the second target network. As can be seen in Figure \ref{fig:celeba_layerwise}, each layer may be understood as having different functionality, i.e. third layer is responsible for the general shape, while the last layer corrects the colors in the image.

\subsection{Super-resolution} \label{sec:superres}

The hypernetwork allows us to describe every image as a continuous function (target network) defined on a unit 2D square. As a result, we can evaluate this function on every grid and upscale the image to any size. In this experiment, we compare the effects of this process with super-resolution approaches. 

To make this approach successful we feed the hypernetwork with low resolution images and evaluate MSE loss on high resolution ones. More precisely, we take the original image of the size $m \times n$, downscale it to $k \times l$ using bicubic interpolation and input it to the hypernetwork. The hypernetwork produces the weights to target network, which defines the functional representation of the input image. To evaluate its quality, we take a grid of the size $m \times n$ on the image returned by target network and compare the values of this grid with pixels intensities of the original image.

Since input images can have different resolutions, we split them into overlapping parts of fixed sizes. In consequence, the value at each coordinate is described by multiple target networks. To produce a single output for every coordinate at test phase, we take the (weighted) average of values returned by all target networks covering this coordinate. This also allows us to smooth the output function. Moreover, we supplied a target network with an additional parameter $\alpha$, which indicated the scaling factor.

To test our approach, we train the model on $800$ examples from \texttt{DIV2K} data set \cite{agustsson2017ntire}. Its performance is evaluated on \texttt{Set5} \cite{bevilacqua2012low}, \texttt{Set14} \cite{zeyde2010single}, \texttt{B100} \cite{martin2001database}, and \texttt{Urban100} \cite{huang2015single}. As a quality measure we use PSNR and SSIM \cite{wang2004image}, which are common score functions applied in super-resolution. Their high values indicate better performance of a model. We consider scale factors $\times 2, \times 3, \times 4$.

\begin{table*}[t]
\centering
\caption{The average PSNR values obtained for a super-resolution task. \label{tab:psnr}}
\begin{tabular}{|c|c|c|c|c|c|}
\hline
\multicolumn{1}{|l|}{\multirow{2}{*}{}} & \multirow{2}{*}{\textbf{Scale}} &
\multirow{2}{*}{\textbf{Bicubic}} & \multirow{2}{*}{\textbf{SRCNN} \cite{SRCNN-2016}} & \multirow{2}{*}{\textbf{RDN} \cite{zhang2018residual}} & \multirow{2}{*}{\textbf{Hypernetwork}} \\
 
\multicolumn{1}{|l|}{} &  &  &  &  &  \\ \hline
\multirow{3}{*}{Set5} & 2x & 33.64 & 36.66 & 38.30 & 36.09 \\ \cline{2-6}
& 3x & 30.41 & 32.75 & 34.78 & 32.85 \\ \cline{2-6}
& 4x & 28.42 & 30.49 & 32.61 & 30.69 \\ \hline
\multirow{3}{*}{Set14} & 2x & 30.33 & 32.45 & 34.10 & 32.30 \\ \cline{2-6}
& 3x & 27.63 & 29.30 & 30.67 & 29.37 \\ \cline{2-6}
& 4x & 26.08 & 27.50 & 28.92 & 27.61 \\ \hline
\multirow{3}{*}{B100} & 2x & 29.48 & 31.36 & 32.40 & 31.11 \\ \cline{2-6}
& 3x & 27.12 & 28.41 & 29.33 & 28.31 \\ \cline{2-6}
& 4x & 25.87 & 26.90 & 27.80 & 26.86 \\ \hline
\multirow{3}{*}{Urban100} & 2x & 26.85 & 29.50 & 33.09 & 29.43 \\ \cline{2-6}
& 3x & 24.43 & 26.24 & 29.00 & 26.26 \\ \cline{2-6}
& 4x & 23.11 & 24.52 & 26.82 & 24.56 \\ \hline
 
\end{tabular}
\end{table*}

\begin{table*}[t]
\centering
\caption{The average SSIM values  obtained for a super-resolution task. \label{tab:ssim}}
\begin{tabular}{|c|c|c|c|c|c|}
\hline
\multicolumn{1}{|l|}{\multirow{2}{*}{}} & \multirow{2}{*}{\textbf{Scale}} &
\multirow{2}{*}{\textbf{Bicubic}} & \multirow{2}{*}{\textbf{SRCNN} \cite{SRCNN-2016}} & \multirow{2}{*}{\textbf{RDN} \cite{zhang2018residual}} & \multirow{2}{*}{\textbf{Hypernetwork}} \\
 
\multicolumn{1}{|l|}{} &  &  &  &  &  \\ \hline
\multirow{3}{*}{Set5} & 2x &  0.930 & 0.9542 & 0.9616 & 0.9505 \\ \cline{2-6}
& 3x & 0.869 & 0.909 & 0.930 & 0.910 \\ \cline{2-6}
& 4x & 0.812 & 0.863 & 0.900 & 0.869 \\ \hline
\multirow{3}{*}{Set14} & 2x & 0.869 & 0.907 & 0.922 & 0.900 \\ \cline{2-6}
& 3x &  0.775 & 0.822 & 0.848 & 0.816 \\ \cline{2-6}
& 4x & 0.703 & 0.751 & 0.789 & 0.751 \\ \hline
\multirow{3}{*}{B100} & 2x & 0.843 & 0.888 & 0.902 & 0.879 \\ \cline{2-6}
& 3x & 0.738 & 0.786 & 0.811 & 0.778 \\ \cline{2-6}
& 4x & 0.666 & 0.710 & 0.743 & 0.706 \\ \hline
\multirow{3}{*}{Urban100} & 2x & 0.839 & 0.895 & 0.937 & 0.891 \\ \cline{2-6}
& 3x & 0.733 & 0.799 & 0.868 & 0.798 \\ \cline{2-6}
& 4x & 0.656 & 0.722 & 0.807 & 0.723 \\ \hline
 
\end{tabular}
\end{table*}

As a baseline, we use bicubic interpolation. Moreover, we compare our approach with SRCNN \cite{SRCNN-2016}, which was a state-of-the-art method in 2016. We also include a recent state-of-the-art method -- RDN \cite{zhang2018residual}. Our goal is to train a single hypernetwork model to generate images at various scales. This is a more general solution than typical super-resolution approaches, where every model is responsible for upscaling the image to only one resolution. Therefore, it is expected that both SRCNN and RDN will perform better than our method.

The results presented in Tables \ref{tab:psnr} and \ref{tab:ssim} demonstrate that our model gives significantly better performance than baseline bicubic interpolation (see also Figure \ref{fig:img1} for sample result). Surprisingly, a single hypernetwork trained on all scales achieves a comparable performance to SRCNN, which created a separate model for each scale factor. It shows high potential of our model. Nevertheless, our model was not able to obtain scores as high as recent state-of-the-art super-resolution methods. It might be caused by insufficient architecture of hypernetwork. In our opinion, designing similar architecture of hypernetwork to RDN should lead to comparable performance. The main advantage of our approach is its generality -- we train a single model for various scale factors.

\section{Conclusion}

In this work, we have presented an extension of hypernetworks mechanism, which makes it possible to create functional representations of images. Due to the continuity of the representation, we were able to upscale images to any resolution, while obtaining results comparable to those achieved by specialized super-resolution models. We also observed that the hypernetwork transforms a manifold of images to a more compact space with a convenient topology. In particular, we were able to traverse linearly from one image to another (using the weights of their target networks) without falling out from the true data distribution. 

In future work, we will investigate other applications of the proposed hypernetwork-based functional image representation. One natural direction of further research is testing other image restoration tasks, such as deblurring, denosing and deblocking. We will also investigate the potential of hypernetworks in image inpainting. Furthermore, we are going to ensure better continuity of the learned representation, by using derivative-based methods such as Sobolev Training \cite{czarnecki2017sobolev}, and based on the improved representation, we will work on implementing continuous filters for convolutional neural networks.

Another research direction is to further explore the geometry of the target network space. For example, we will try to use the distance in the weights space between two target network-based representations as an alternative to the typical mean squared error loss used in many deep learning models, e.g. autoencoders or generative models.

\section*{Acknowledgements}

This work was partially supported by the National Science Centre (Poland) grant no. 2018/31/B/ST6/00993.

%
\bibliographystyle{splncs04}
\bibliography{icann19}

\end{document}